\documentclass[10pt,twocolumn,letterpaper]{article}
\usepackage[pagenumbers]{vendor/cvpr}      %
\usepackage{amsmath,amssymb,graphicx,booktabs}
\usepackage{xcolor}
\definecolor{cvprblue}{rgb}{0.21,0.49,0.74}
\usepackage[pagebackref,breaklinks,colorlinks,allcolors=cvprblue]{hyperref}
\usepackage{xurl}
\usepackage{capt-of}
\ifdefined\XeTeXrevision\usepackage[OT1]{fontenc}\fi %

\newif\ifwithrldr \withrldrtrue

\newcommand{\lego}{LEGO\textsuperscript{\textregistered}}  %
\title{BriConstrain: Geometrically Constrained Decoding \\ for Autoregressive Brick Assembly Generation}
\def\confName{CVPR}
\def\confYear{2027}
\author{Nadav Magar$^{1}$ \quad Amir Hertz$^{2}$ \quad Ariel Shamir$^{3}$ \quad Or Patashnik$^{1}$\\
$^{1}$Tel Aviv University \quad $^{2}$Google DeepMind \quad $^{3}$Reichman University}
\begin{document}
\twocolumn[{\renewcommand\twocolumn[1][]{#1}\maketitle\begin{center}
\includegraphics[width=\textwidth]{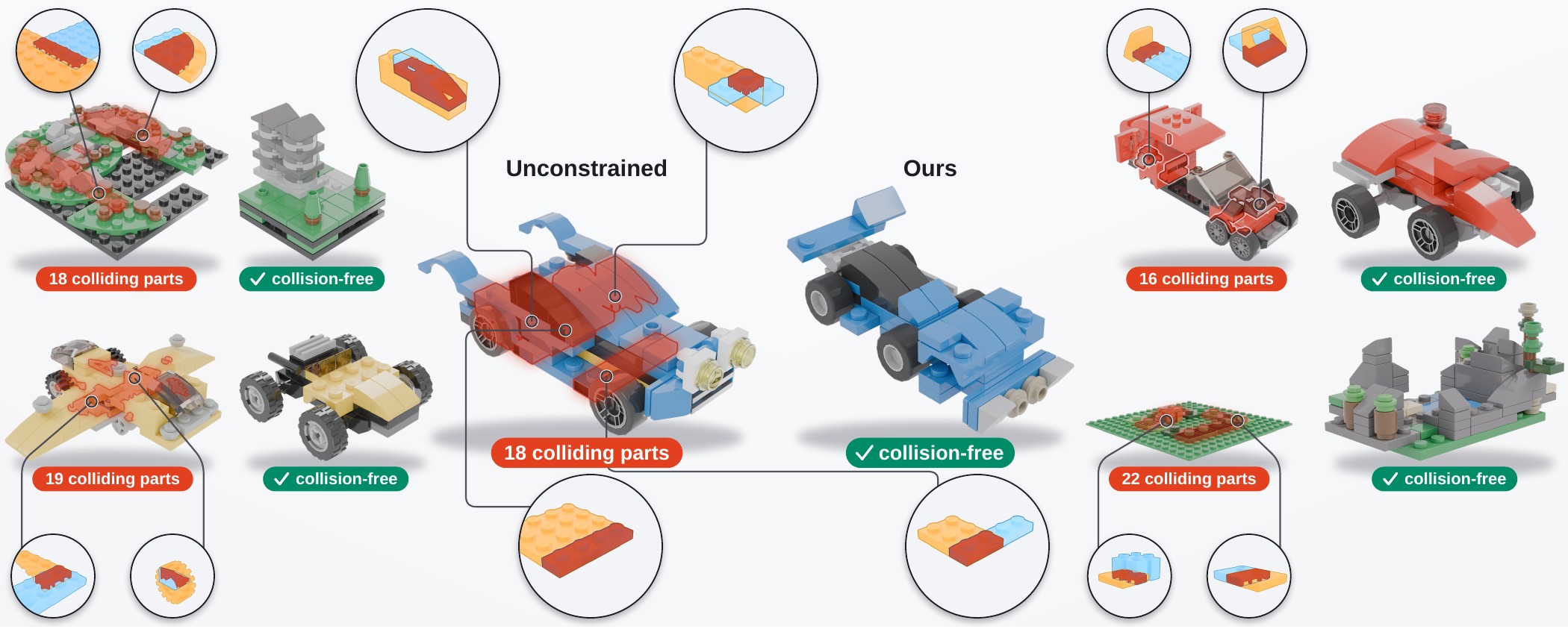}
\captionof{figure}{
    We constrain the output of large language models for \lego generation to enforce geometric validity.
    Each scene shows a build decoded freely (left) and with our method (right) for the same caption, by BrickNet-14B or, bottom left, by our ICLF model. 
    Red marks interpenetrating parts; each inset zooms in on a colliding pair (bricks in blue and orange, the volume both occupy in red). 
}
    \label{fig:teaser}
\end{center}
}]
\begin{abstract}
\par\vspace{-10pt}\noindent  %
Language models can generate \lego{} assemblies from text, but their builds often contain parts that collide or attach to nothing. 
The usual approach of rejecting and resampling invalid placements discards decoded tokens without guaranteeing a valid build.
We present BriConstrain, a constrained decoding method that restricts a pretrained language model's next-token probabilities before sampling, steering it towards valid placements.
Because the validity of a brick placement depends on the non-predetermined build state, 
it cannot be compiled into a fixed grammar and must be checked dynamically.
However, checking the validity of a brick before sampling is challenging as the space of valid placements is combinatorially large and cannot be enumerated exhaustively.
Our key idea is to narrow the search space by constraining \emph{how} parts attach rather than which parts attach to each other.
Concretely, given the model's predictions of the currently placed part, and the previously placed parent it wants to connect to, we constrain the model's choice of how they attach.
This allows us to span the space of part attachments, using a precomputed geometric oracle, and to effectively filter out colliding options. 
With valid attachment candidates in hand, we bias the model's next-token distribution towards the remaining valid candidates.
We evaluate our method on two \lego{} formats and models: BrickNet, which annotates connectivity explicitly, and our proposed Implicit Connectivity LDraw Format (ICLF), which uses numeric values for placements and leaves connectivity implicit.
Our method raises collision-free builds from 31\% to 91\% on BrickNet and from 15\% to 65\% on ICLF,
and outperforms rejection sampling on both formats at about 60\% of its decoding cost.
\end{abstract}
\vspace{-16pt}  %
\vspace{-10pt}
\section{Introduction}
\label{sec:intro}

Designing a \lego{} model is open-ended: the designer selects from tens of thousands of available parts, 
and combines them to create a desired structure.
At the same time, \lego{} construction is heavily constrained by two main geometric requirements: each part must connect through compatible connectors and must avoid collisions with other parts.
Thus, generating valid \lego{} assemblies is challenging for generative models~\cite{tang2026legopuzzlesgoodmllmsmultistep, kulits2026brickbench}, as it requires geometric understanding of the evolving assembly.
In particular, the validity of each new placement depends on the parts already placed, introducing long-range dependencies when the assembly is generated autoregressively.

Recent works utilize large language models for \lego{} generation, addressing geometric validity through two main approaches.
Some methods aim to learn geometric constraints during training, either implicitly from data~\cite{kulits2026bricknet} or through reward-based optimization~\cite{xu2026stable}. 
However, learning these constraints remains challenging, as illustrated by the collisions on the left of each example in \Cref{fig:teaser}.
Other methods check geometric validity during generation and use rollback and rejection sampling to replace invalid placements~\cite{pun2025brickgpt,Ge2025LEGOMakerAI,ni2026brickanything}.
These methods incur additional computational cost by generating and discarding invalid placements, and may fail even when valid alternatives exist.

In this work we ask: Can large language models trained to generate \lego{} assemblies be guided at inference time to produce geometrically valid builds, while preserving their visual fidelity?
Our key idea is to \textit{constrain} the model's next-token distribution \textit{during decoding}, preventing invalid placements from being sampled in the first place.

This constrained decoding approach presents two main challenges.
First, the set of valid placements depends on the non-predetermined build state, requiring geometric constraints to be evaluated \emph{dynamically} during generation. 
Second, constraining the next-token distribution requires identifying valid placements \emph{before} they are generated,
however the space of placements is vast, with continuous degrees of freedom, making exhaustive enumeration infeasible.

We address these challenges in three steps. 
First, we let the model select the next part and the previously placed part to which it attaches, and focus on constraining \emph{how} they connect. 
This reduces the space of possible placements to a manageable set of relative attachments. 
To enable efficient decoding, we introduce a \textit{geometric oracle}, a precomputed table of possible attachments for each pair of part types. 
Second, during decoding, we dynamically filter these candidates based on collisions and connector availability. 
Third, we apply the constraints by biasing the model's next-token distribution towards valid candidates.

We evaluate our method on two autoregressive \lego{} generation models: the pretrained BrickNet model~\cite{kulits2026bricknet}, which uses \emph{explicit} connectivity, and a model we train with an \emph{implicit} connectivity representation (Sec.~\ref{sec:iclf}). 
Our method reduces their invalid placements by 98\% and 79\% respectively. 
Visual similarity to the reference builds is essentially unchanged on BrickNet and drops slightly on our model (Tab.~\ref{tab:bn_operator_ablation}, Fig.~\ref{fig:qual}).
Compared to rejection sampling with eight attempts, our method produces fewer colliding parts, at $\sim$60\% of the decoding cost on both formats (Fig.~\ref{fig:rejection_cost}).

\section{Related Work}
\label{sec:related_work}
\paragraph{Brick Assembly Generation.}
Early works on generating brick assemblies focus on grid-aligned builds consisting of rectangular pieces.
These methods convert a target shape into bricks by voxel merging and structural repair~\cite{testuz2013brick},
or learn to place bricks sequentially~\cite{thompson2020buildinglegousingdeep,chung2021brickbybrick},
all with a handful of brick types on a grid.
More recently, large language models became popular for autoregressive brick assembly generation.
BrickGPT~\cite{pun2025brickgpt} trains a text-conditioned LLM on Legolized~\cite{Luo2015Legolization} models, over a dictionary of 8 rectangular pieces.
\citet{Ge2025LEGOMakerAI} use a bigger part dictionary and a procedurally generated dataset of building facade assemblies to train a domain-specific image-conditioned LLM.
BrickNet~\cite{kulits2026bricknet} fine-tunes a language model on a large curated set of human-designed LDraw assemblies with diverse parts and connector types.
We target the complete LDraw setting, using the BrickNet dataset and their model, while also showing results on our implicit connectivity format.

\paragraph{Geometric Validity in Brick Generation.}
Prior generators make builds valid either by shaping what the model learns or by checking what it produces.
On the learning side, BrickNet~\cite{kulits2026bricknet} writes every attachment as an explicit pair of connectors and trains on collision-filtered builds, while \citet{xu2026stable} go further and reward collision avoidance and connectivity with reinforcement learning.
Training this way makes invalid placements rarer, but nothing stops the model from sampling one.
On the checking side, a brick is validated only after it is sampled, and resampled or rolled back when it fails~\cite{pun2025brickgpt,Ge2025LEGOMakerAI,ni2026brickanything}.
Every failure then costs another round of decoding: with rollback, BrickGPT takes 895\,s per build, against 58\,s for the RL-trained model of \citet{xu2026stable}.
We move the check to before sampling, so that the model only ever chooses among valid placements.
\citet{ahn2024brecs} also take this approach, masking occupied cells before each brick, but without a language model and with only one or two brick types on a voxel grid.
In contrast, our method restricts pretrained language models over the full LDraw part catalog, which also requires spanning the candidate placements, filtering them efficiently, and biasing the model so that it retains its generative prior.

\paragraph{Constrained LLM Decoding.}
Constrained decoding restricts a language model's next-token distribution at each step, masking the logits of tokens that would make the output invalid.
Commonly, it is applied on simple static constraints which are predetermined and can be compiled into token masks ahead of generation.  
Examples include: formal grammar constraints~\cite{willard2023outlines,dong2025xgrammar}, well-formed mesh generation~\cite{nash2020polygen} and valid sketch-and-extrude sequences in CAD~\cite{wang2026cadtokenizer}.
Dynamic constraints, on the other hand, depend on the generated prefix, cannot be pre-compiled and must be evaluated during generation by querying a checker on the current state prefix.
Examples include checking whether a partial program parses~\cite{scholak2021picard}, executes~\cite{wang2018executionguided}, type-checks~\cite{poesia2022synchromesh,mundler2025type}, or whether a robot action is safe~\cite{kapoor2025safedec}.
Geometric \lego{} validity is dynamic in this sense, since whether a placement is valid depends on every part placed before it.
It also adds a second difficulty: a placement spans several tokens and its validity is determined by all of them, so it has to be checked speculatively ahead of decoding.

\section{Preliminaries}
\label{sec:prelim}
\paragraph{LDraw.}
LDraw~\cite{ldraw} is a community-maintained standard and part library for digital \lego{} modeling, with over 24,000 parts that each replicate a real piece.
An assembly is a plain-text file in which each line places one part instance by its part ID, color, absolute position and rotation matrix
(see Fig.~\ref{fig:appendix_one_build} in the supplementary).
The format stores geometry only, so connections must be inferred from connector metadata and compatibility rules~\cite{ldcadshadow}, and invalid builds are allowed.
The use of non-natural language IDs, absolute coordinates and no connectivity hints makes the format difficult for LLMs to learn.

Our method can be applied only to formats where parent-child relations are defined and handled by the model, constraining how parts attach. 
We demonstrate our approach on two different formats and models --- our own Implicit Connectivity LDraw Format (ICLF) and the BrickNet~\cite{kulits2026bricknet} explicit connectivity format.
These reflect different design choices for representing connectivity, and the effect of this choice is studied in Sec.~\ref{sec:formats}.

\begin{figure*}[t]
\centering
\includegraphics[width=\textwidth]{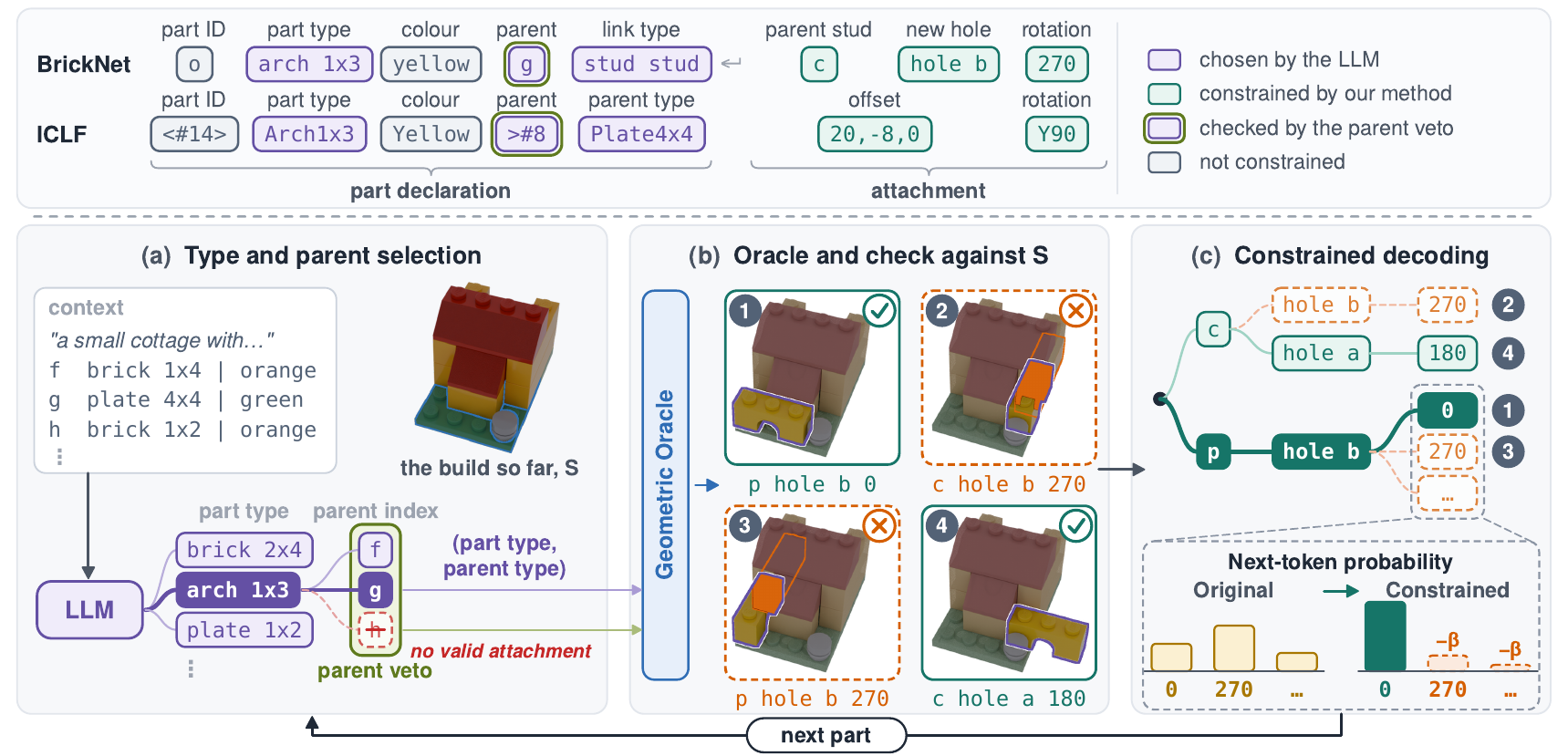}
\caption{
  Constrained decoding of one part.
  Top: how one part is written in BrickNet's format and in ICLF; colors mark which fields the model chooses freely and which our method constrains.
  (a)~Given the caption and the parts built so far, the model selects the next part type and its parent; parents without any valid attachment are vetoed (olive box; the vetoed parent is shown in red, Sec.~\ref{sec:parent_veto}).
  (b)~The geometric oracle lists the ways the part can attach to its parent (Sec.~\ref{sec:oracle}); each candidate is placed in the build so far, $S$, and checked for collisions.
  (c)~The candidates, written as tokens, form a tree; tokens that lead only to colliding candidates
  are penalized by $-\beta$ before sampling, and removed in the hard-mask case $\beta\to\infty$ (Sec.~\ref{sec:token_constraints}). 
  In the tree, \ldots{} stands for the continuations the oracle does not offer. The dashed box zooms in on the last decoded token: its original next-token probabilities and the constrained ones, in which the colliding 270 and the continuations the oracle does not offer are lowered by $\beta$.
  The decoded part joins the build and the loop repeats.
}
\label{fig:method_overview}
\end{figure*}

\paragraph{BrickNet.}
\label{sec:bricknet}  %
We briefly summarize the BrickNet format here for clarity, for full details see the original paper~\cite{kulits2026bricknet}.
BrickNet annotates LDraw parts with typed, named connectors, each leaving few degrees of freedom, such as the yaw of a stud mate.
A build is serialized as a spanning tree of its connection graph: each step declares a new part and its color, followed by a connectivity edge. 
The connectivity edge names a previously placed parent, the connector family, the two connector IDs across the parts and the remaining pose parameters
(Fig.~\ref{fig:method_overview}, top).
A compiler derives the exact offset and transform from this format and outputs regular LDraw files.
Connectivity is thus explicit.

BrickNet released text-conditioned models of up to 14B parameters, which we use without retraining, and a mesh-based collision checker.
To check candidate placements during decoding, we use their meshes and implement an efficient collision checker on GPU, batching candidate poses against the current build (Sec.~\ref{app:mesh_checker} in the supplementary).

\paragraph{Implicit Connectivity LDraw Format (ICLF).}
\label{sec:iclf}
We propose a relative LDraw format with implicit connectivity and numeric placement values.
ICLF places each part relative to its parent: a part declaration is a single line with the part name, color and parent, followed by the offset and the rotation (Fig.~\ref{fig:method_overview}, top).
The offset is in LDraw units in the parent's frame, and the rotation is one of the 24 axis-aligned rotations (\eg \texttt{Y90}), with axis--angle or matrix notation for the rest.
No connectors are named, so connectivity is implicit, following only from the geometry, and the model commits to an attachment through the offset coordinates and the rotation.
We fine-tune Qwen3.6-27B~\cite{qwen36} on captioned ICLF builds, including BrickNet's training builds converted to ICLF (Sec.~\ref{sec:models}).
For the full implementation details refer to Sec.~\ref{app:rldr_details} in the supplementary.

\section{Method}
\label{sec:method}

Our method enforces geometric validity on a pretrained language model's output by restricting the model's next-token distribution to known valid placements.
Sec.~\ref{sec:oracle} describes how we build and use the geometric oracle, which lists the candidate placements of a new part.
Sec.~\ref{sec:token_constraints} describes how we filter these candidates against the assembly and turn the remaining candidates into token constraints. 
Finally, Sec.~\ref{sec:formats} explains how our method applies to both implicit and explicit representations of connectivity information.  

\subsection{Geometric Placement Oracle}
\label{sec:oracle}

A placement is serialized to a series of several tokens, and all are required to determine whether it collides or not.
Therefore, to prevent sampling an invalid placement, we must first list all possible placements, and check which of them are valid in the context of the current build, ahead of sampling an attachment.
Listing and checking every placement of every part type in the \lego{} catalog is intractable.

Instead, we let the model choose the part type and the previously placed parent part, to which it attaches, and constrain only how they attach.
The set of possible attachments depends only on the part types of the currently placed brick and its parent and, crucially, not on the rest of the build, so it can be precomputed.
We build and store this static map, which we call the \emph{geometric oracle} (panel b in Fig.~\ref{fig:method_overview}): 
for a given part type $t$ and a parent of type $t_p$ the oracle returns a finite set $C(t,t_p)$ of candidate attachments.
Each attachment consists of the connector-to-connector pairing across the parts and its pose parameters.

We build the oracle from two complementary sources:
The first enumerates all the pairs of part types, and for each pair every compatible connector pairing across the part types. 
For each such connector pairing, we sample the connection's pose free variables from a coarse grid of connector specific values (full details in Sec.~\ref{sec:oracle_coverage} in the supplementary).
We measure how expressive the geometric oracle is by its coverage of the attachments used in our validation set.
Enumeration covers nearly all stud attachments (98\%) but only 15\% of axle and 43\% of hinge attachments, whose angles and slides are continuous.

We therefore add \emph{empirical} attachments from the training builds, which supply common off-grid attachments that a grid cannot reach.
Together the two sources raise axle and hinge coverage to 68\% and 92\%, and cover 96\% of the attachments in the validation set (Table~\ref{tab:oracle_coverage} in the supplementary).
An attachment missing from both sources can never be proposed by the oracle.
\subsection{Geometric Filters as Token Constraints}
\label{sec:token_constraints}
\definecolor{tabfirst}{rgb}{0.96, 0.77, 0.77}
\definecolor{tabsecond}{rgb}{0.99, 0.85, 0.66}
\definecolor{tabthird}{rgb}{1.00, 0.97, 0.70}
\definecolor{ownchk}{gray}{0.82}
\providecommand{\cfirst}[1]{\colorbox{tabfirst}{\makebox[0.66cm][c]{#1}}}
\providecommand{\csecond}[1]{\colorbox{tabsecond}{\makebox[0.66cm][c]{#1}}}
\providecommand{\cthird}[1]{\colorbox{tabthird}{\makebox[0.66cm][c]{#1}}}
\providecommand{\cnone}[1]{\makebox[0.66cm][c]{#1}}
\begin{table*}[t]
\centering\small  %
\setlength{\tabcolsep}{2pt}\setlength{\fboxsep}{1pt}
\begin{tabular}{lrccccccccc}
\toprule
 & & \multicolumn{3}{c}{BrickNet checker (\%)} & \multicolumn{3}{c}{ColSAT (\%)} & \multicolumn{3}{c}{Visual proxies}\\
\cmidrule(lr){3-5}\cmidrule(lr){6-8}\cmidrule(lr){9-11}
Arm & Parts & CF\,$\uparrow$ & Part\,inv.\,$\downarrow$ & Build\,inv.\,$\downarrow$ & CF\,$\uparrow$ & Part\,inv.\,$\downarrow$ & Build\,inv.\,$\downarrow$ & SigLIP\,$\uparrow$ & DINO\,$\uparrow$ & VQA\,$\uparrow$\\
\midrule
\multicolumn{11}{l}{\textit{BrickNet-14B, BN-val-512 (2048 attempts per arm)}}\\
B0 Freeform & 30.7 & 30.9 & 23.7 & 16.5 & 25.7 & 23.6 & 18.1 & \colorbox{tabfirst}{0.152} & \colorbox{tabsecond}{0.630} & \colorbox{tabfirst}{0.264}\\
B1 + grammar \& catalog & 30.6 & 29.1 & 24.8 & 17.7 & 24.7 & 24.2 & 18.9 & \colorbox{tabsecond}{0.151} & \colorbox{tabfirst}{0.632} & \colorbox{tabthird}{0.257}\\
B2 + collision-aware mates & 33.3 & 32.7 & 19.8 & 14.2 & 27.6 & 20.2 & 16.3 & \colorbox{tabsecond}{0.151} & \colorbox{tabsecond}{0.630} & 0.255\\
B3 + collision-aware placements & 36.1 & \colorbox{tabthird}{47.4} & \colorbox{tabthird}{7.4} & \colorbox{tabthird}{6.0} & \colorbox{tabthird}{39.1} & \colorbox{tabthird}{9.8} & \colorbox{tabthird}{9.5} & \colorbox{tabthird}{0.150} & 0.626 & \colorbox{tabsecond}{0.261}\\
B4 + parent veto (ours) & 35.4 & \colorbox{tabfirst}{90.8} & \colorbox{tabfirst}{0.5} & \colorbox{tabfirst}{1.0} & \colorbox{tabfirst}{64.1} & \colorbox{tabfirst}{4.2} & \colorbox{tabfirst}{5.7} & \colorbox{tabthird}{0.150} & \colorbox{tabthird}{0.627} & \colorbox{tabthird}{0.257}\\
Rejection sampling, $N{=}8$ & 30.7 & \colorbox{tabsecond}{74.3} & \colorbox{tabsecond}{2.7} & \colorbox{tabsecond}{1.7} & \colorbox{tabsecond}{50.8} & \colorbox{tabsecond}{6.9} & \colorbox{tabsecond}{7.5} & \colorbox{tabthird}{0.150} & \colorbox{tabfirst}{0.632} & 0.254\\
\ifwithrldr
\midrule
\multicolumn{11}{l}{\textit{ICLF, 427 non-minifig BN-val captions (1708 attempts per arm)}}\\
I0 Freeform & 34.1 & 14.6 & 27.3 & 25.2 & 16.7 & 22.2 & 20.4 & \colorbox{tabsecond}{0.145} & \colorbox{tabfirst}{0.600} & \colorbox{tabsecond}{0.296}\\
I1 + grammar \& catalog & 34.9 & 7.4 & 34.1 & 32.2 & 10.7 & 27.4 & 25.5 & \colorbox{tabsecond}{0.145} & \colorbox{tabthird}{0.596} & 0.279\\
I2 + collision-aware placements & 35.4 & \colorbox{tabthird}{31.4} & \colorbox{tabthird}{12.1} & \colorbox{tabthird}{11.9} & \colorbox{tabsecond}{31.7} & \colorbox{tabthird}{11.0} & \colorbox{tabthird}{10.8} & \colorbox{tabthird}{0.144} & \colorbox{tabsecond}{0.597} & \colorbox{tabthird}{0.292}\\
I3 + parent veto (ours) & 31.2 & \colorbox{tabfirst}{65.2} & \colorbox{tabfirst}{5.8} & \colorbox{tabfirst}{6.4} & \colorbox{tabfirst}{58.7} & \colorbox{tabfirst}{6.0} & \colorbox{tabfirst}{6.5} & 0.142 & 0.593 & 0.262\\
Rejection sampling, $N{=}8$ & 33.0 & \colorbox{tabsecond}{32.8} & \colorbox{tabsecond}{11.1} & \colorbox{tabsecond}{10.2} & \colorbox{tabthird}{29.7} & \colorbox{tabsecond}{10.5} & \colorbox{tabsecond}{9.9} & \colorbox{tabfirst}{0.149} & \colorbox{tabsecond}{0.597} & \colorbox{tabfirst}{0.340}\\
\fi
\bottomrule
\end{tabular}
\caption{
  Constraint ablation.
  Each row adds one constraint to the row above (Sec.~\ref{sec:token_constraints}); the last row of each block is rejection sampling, which re-draws each colliding part up to $N{=}8$ times (Sec.~\ref{sec:rejection}).
  Top: BrickNet-14B on BN-val-512; bottom: our ICLF model on 427 BN-val captions; four samples per caption.
  Parts: mean number of parts per parsable build; the other metrics are defined in Sec.~\ref{sec:metrics}.
  Colors mark the three best values in each column, per block.
}
\label{tab:bn_operator_ablation}
\end{table*}

The geometric oracle lists how a part can attach to its parent, but not which of these attachments fit the current build $S$. 
We therefore filter the oracle candidates against $S$, rejecting those that collide with it.
For a part of type $t$ and a parent of type $t_p$, we denote the subset of valid candidates in $S$ as
\begin{equation}
  F_p(S)=\{c\in C(t,t_p) : c \text{ is valid in }S\}
\end{equation}
where validity is determined by our efficient GPU collision checker implementation (Sec.~\ref{app:mesh_checker} in the supplementary).

Since the model writes an attachment token by token, the list of valid attachments must be converted to next-token constraints. 
We tokenize the remaining candidates and build a prefix tree on them, so the model chooses a logical attachment incrementally token by token (each token is a node of the tree).
The process is shown in Figure~\ref{fig:method_overview} panel (c).
We examine how the choice of format language affects the shape of the prefix tree in Sec.~\ref{sec:formats}.
Formally, at decoding step $i$, with $a_{<i}$ the attachment tokens decoded so far, the set of valid next tokens is given by
\begin{equation}
  A_i=\{v:\exists c\in F_p(S),\ a_{<i}v\preceq s(c)\},
\end{equation}
where $s(c)$ denotes the tokenization of attachment $c$ and $\preceq$ denotes the prefix relation.

Beneath the geometric constraint, we restrict the model to the format's grammar, which removes tokens that would make the serialization unparsable or name a non-existent part.
Let $z_i$ be the model's next-token logits given the tokens $y_{<i}$ decoded so far. 
We bias towards valid attachments by penalizing tokens outside $A_i$:
\begin{equation}
  p_\beta(v\mid y_{<i})=\operatorname{softmax}\bigl(z_i(v)+g_i(v)-\beta\,b_i(v)\bigr),
  \label{eq:soft_reference}
\end{equation}
where $g_i(v)$ is $0$ for tokens the grammar allows and $-\infty$ otherwise, $b_i(v)=\mathbf{1}[A_i\neq\emptyset,\ v\notin A_i]$ indicates tokens outside $A_i$, and $\beta\ge 0$ is a hyperparameter.
Note that $\beta$ sets the threshold for how confident the model must be to overrule the oracle.
As $\beta\to\infty$ the model is restricted to oracle-valid continuations, and at $\beta=0$ no geometric constraint is imposed.
A finite $\beta$ is appealing because a hard restriction may force the model to make a geometrically valid, but aesthetically poor attachment choice when only few valid candidates exist (oracle coverage is incomplete).
If no candidate survives, or if the model overrules the oracle, the rest of the attachment is decoded with the grammar alone.
In Sec.~\ref{sec:hard_vs_soft} we sweep $\beta$ to test whether letting the model overrule the oracle helps, and default to the hard mask ($\beta\to\infty$) unless otherwise noted.

\paragraph{Parent veto.}
\label{sec:parent_veto}
The placement filter acts only after the parent is chosen, so a parent with no surviving candidate forces the grammar-only fallback.
We prevent choosing a parent that will result in grammar-only fallback by adding a constraint on the parent token: a parent must have at least one valid attachment, i.e.\ $F_p(S)\neq \emptyset$, falling back to the grammar if no such parent exists.
\begin{figure*}[t]
\centering
\includegraphics[width=\textwidth]{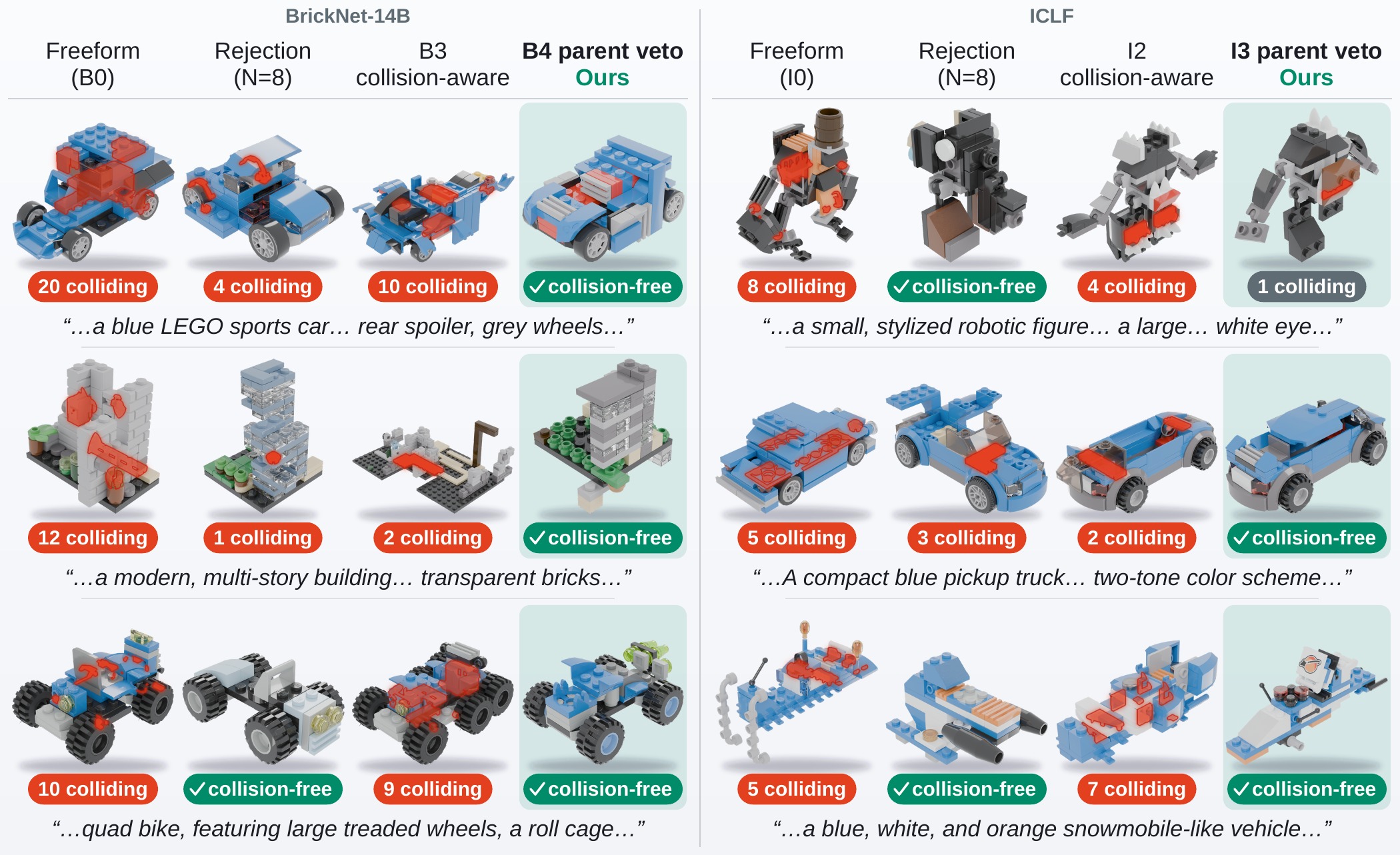}
\caption{
Qualitative comparison on BrickNet validation captions. 
Each row shows BrickNet-14B (left group) and our ICLF model (right group). The quoted excerpts are the literal model inputs.
Columns: freeform decoding (B0/I0), rejection sampling with $N{=}8$ tries per part, the placement filter alone (B3/I2),
 and our full method with the parent veto (B4/I3). 
Colliding parts are tinted red and counted in the pills.
}
\label{fig:qual}
\end{figure*}
\subsection{Format and Prefix-Tree Shape}
\label{sec:formats}

The oracle and the filtering operations described in previous sections are independent of the assembly format. 
The format only affects how logical attachments are tokenized into a prefix tree.
The shape of the tree determines how the attachment choices spread over the decoded tokens, and thus which choice the model makes at each token.
To measure this, we sample part-type pairs from BrickNet's validation builds, weighted by how often they occur, write the same candidate placements in both formats, tokenize them with each model's tokenizer, and count the children per node at each level of the resulting trees (Sec.~\ref{app:rldr_trie} and Table~\ref{tab:trie_shape} in the supplementary).
BrickNet's tree is wide and shallow: an attachment takes 9.6 tokens, and the choice of candidates is mostly made at the two connector tokens, which split into about 16 and 5 branches on average.
ICLF's tree is deeper and narrower: an attachment takes 12.7 tokens, and apart from two levels with about 6 and 2 branches, every node has fewer than two children, so the choice spreads over the digits of the offset and the rotation.
Our method successfully reduces invalid placements on both ICLF and the BrickNet formats (Sec.~\ref{sec:operator_ablation}).
Since the ICLF and BrickNet models differ in architecture and training data (Sec.~\ref{sec:experiments}), 
we compare constraints within each model and not across them.

\vspace{-8pt}  %
\section{Experimental Setup}
\label{sec:experiments}

\subsection{Models and Datasets}
\paragraph{Models.}
\label{sec:models}
Our ICLF model is based on Qwen3.6-27B~\cite{qwen36}, fine-tuned with a rank-512 LoRA (training details in Sec.~\ref{app:rldr_training} in the supplementary).
We train it on BrickNet's training builds converted to ICLF.
We remove \lego{} minifigures from the builds, as their parts and orientations follow an irregular distribution and are mostly decorative additions to the builds.
We also use BrickNet-14B~\cite{kulits2026bricknet}, which is based on Qwen3~\cite{yang2025qwen3}, without retraining.
As the two models differ in data and representation, we compare constraints within each model.

\paragraph{Evaluation data.}
We evaluate on BrickNet's validation set of 512 captions (BN-val-512), held out from training, with four samples per caption.
For the ICLF model we omit minifigure-centered builds, leaving 427 captions.

\subsection{Metrics}
\label{sec:metrics}
\label{sec:validators}
\paragraph{Structural validity.}
We count a part as invalid if it collides with an earlier part.
We report the percentage of collision-free builds (\textit{CF}), defined as builds with no detected collisions, and two invalidity metrics: \emph{Part inv.}, the fraction of invalid parts across all builds, and \emph{Build inv.}, the average over all builds of the fraction of invalid parts per build. These measures are defined as:

\begin{equation}
  P_{\mathrm{part}}=\frac{\sum_b c_b}{\sum_b n_b},\qquad
  P_{\mathrm{build}}=\frac{1}{B}\sum_{b=1}^{B}\frac{c_b}{n_b},
\end{equation}

where $B$ is the number of builds and for each build $b$, $n_b$ is the number of parts and $c_b$ is the number of invalid parts. 

We detect collisions with two checkers:
the first is BrickNet's mesh checker, which our decoder also queries (Sec.~\ref{app:mesh_checker} in the supplementary).
The second, \emph{ColSAT}, follows the check human designers rely on: it tests collisions using oriented collision boxes from BrickLink Studio~\cite{bricklinkstudio}, an LDraw editor, \ie the collisions a designer sees flagged while building (Sec.~\ref{app:colsat_axes} in the supplementary).
ColSAT uses a different collision test from the one used during decoding, helping identify cases where the decoder exploits inaccuracies in the BrickNet checker.
Neither checker is exact, as connectors can touch without overlapping and LDraw meshes are not watertight, potentially causing both false positive and false negative results in both checkers.
\paragraph{Visual agreement.}
We measure prompt adherence using SigLIP2~\cite{tschannen2025siglip2} image--caption similarity and VQAScore~\cite{lin2024evaluating}, and resemblance to the reference build with DINOv2~\cite{oquab2024dinov2} image--image similarity.
Each metric is averaged over front and back renders, since a build may be recognizable from only one side.

\section{Results}
\label{sec:results}

\subsection{Comparisons}
\label{sec:rejection}

We compare our method against two baselines for each model: unconstrained decoding and rejection sampling~\cite{pun2025brickgpt,Ge2025LEGOMakerAI,ni2026brickanything}.
Rejection sampling is a natural approach to enforcing structural validity, but it relies on the model proposing valid placements and requires decoding each unsuccessful attempt.
In contrast, our method filters invalid placements before sampling.
Table~\ref{tab:bn_operator_ablation} compares their validity and visual fidelity, and Fig.~\ref{fig:qual} shows qualitative examples.

\begin{figure}[t]
\centering
\ifwithrldr\includegraphics[width=\columnwidth]{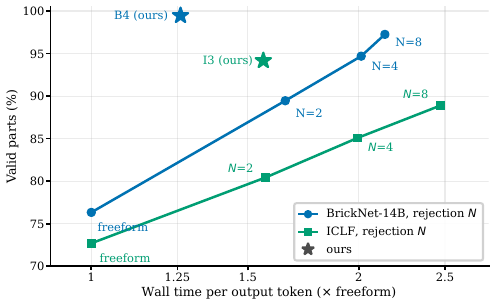}\else\includegraphics[width=\columnwidth]{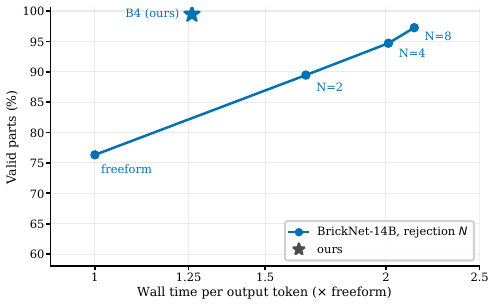}\fi
\caption{
Validity against cost. Share of valid parts against wall time per output token (total decoding time, including rejected attempts, per token kept in the final build) relative to freeform decoding. 
We compare rejection sampling with $N$ tries (circles and squares) with our method (stars) on both BrickNet and ICLF (names match Table~\ref{tab:bn_operator_ablation}). 
Our method produces more valid parts than any rejection budget, at a lower cost.
}
\label{fig:rejection_cost}
\end{figure}
\paragraph{Validity.}
Under unconstrained decoding, only 30.9\% of the assemblies generated by BrickNet-14B are collision-free (Table~\ref{tab:bn_operator_ablation}).
Rejection sampling improves the percentage of collision-free builds from 30.9\% to 49.6\% with up to two attempts per part, and to 74.3\% with eight attempts (Table~\ref{tab:rejection} in the supplementary).
Furthermore, rejection sampling cannot resolve an invalid parent choice: if the selected parent has no valid attachment for the new part, repeated sampling may fail to find a valid placement.

The parent veto removes such parents before the part is written (Sec.~\ref{sec:token_constraints}), and our full method achieves 90.8\% collision-free builds, with only 0.5\% of parts invalid compared to 2.7\% for rejection sampling with $N{=}8$.
Our method also outperforms rejection sampling and unconstrained decoding under ColSAT, an independent collision checker not used during decoding. This suggests that the improvement is not specific to the checker used during generation.
On ICLF, our method (I3) outperforms rejection sampling with $N{=}8$, doubling collision-free builds (65.2\% vs.\ 32.8\%) and halving invalid parts (5.8\% vs.\ 11.1\%).

\paragraph{Cost.}
Every rejected attempt costs additional decoding, so the cost of rejection sampling grows with the number of attempts (Fig.~\ref{fig:rejection_cost}).
We measure cost as wall time per output token: the total decoding time, including rejected attempts, divided by the number of tokens in the final build, since build length varies across methods and prompts.
On BrickNet-14B, rejection sampling with two, four, and eight attempts takes 1.65$\times$, 2.01$\times$, and 2.14$\times$ the time per output token of unconstrained decoding.
Our method (B4) needs no additional model calls; its only overhead is checking the candidate placements on the GPU (Sec.~\ref{sec:bricknet}), which brings its cost to 1.26$\times$, while it also leaves fewer invalid parts.
On ICLF, our method (I3) takes 1.56$\times$ the time per output token of unconstrained decoding, compared to 2.47$\times$ for rejection sampling with up to eight attempts per part.

\paragraph{Fidelity.}
On BrickNet-14B, both our method (B4) and rejection sampling largely preserve visual fidelity compared to unconstrained decoding, with similar DINO similarities and VQA scores (Table~\ref{tab:bn_operator_ablation}).
On ICLF, our method (I3) keeps DINO similarity close to unconstrained decoding and rejection sampling (0.593 vs.\ 0.600 and 0.597) but lowers the VQA score (0.262 vs.\ 0.296 and 0.340).
Figure~\ref{fig:qual} shows examples from both models, illustrating how our method reduces collisions while largely preserving the appearance of the builds.

\begin{figure}[t]
\centering
\ifwithrldr\includegraphics[width=\columnwidth]{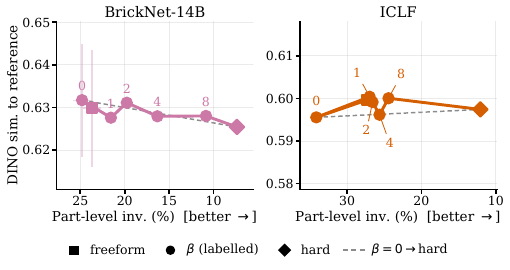}\else\includegraphics[width=\columnwidth]{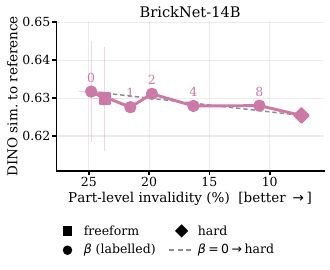}\fi  %
\caption{
  Hard versus soft constraints on BrickNet-14B (left) and ICLF (right). 
  Each curve sweeps the placement penalty from freeform decoding (square) through $\beta=0, 1, 2, 4,8$ (circles) to the hard mask (diamond).
}
\label{fig:soft_pareto}
\end{figure}
\subsection{Geometric Constraint Ablations}
\label{sec:operator_ablation}

We conduct an ablation study to evaluate the contribution of each constraint in our decoding method.
Table~\ref{tab:bn_operator_ablation} progressively adds constraints, from unconstrained decoding to our full method (Sec.~\ref{sec:token_constraints}).
We first apply grammar and catalog constraints (B1, I1), which improve the percentage of parsable outputs on BrickNet-14B from 94.2\% to 99.5\% (Table~\ref{tab:bn_operator_cost} in the supplementary), but do not reduce geometric invalidity (Table~\ref{tab:bn_operator_ablation}).

\paragraph{Connector Constraints vs. Full Placement Constraints.}
For BrickNet, we first evaluate constraining only the connector pair (B2), while leaving the remaining pose parameters unconstrained.
This reduces invalid parts only from 23.7\% to 19.8\%, since the model can still select an invalid pose, such as a rotation about the connector axis.
In contrast, ICLF does not explicitly represent connector pairs, so this ablation is not applicable.
Constraining the complete placement (B3 for BrickNet and I2 for ICLF), including filtering occupied connectors, substantially reduces invalid parts to 7.4\% and 12.1\%, respectively.
On BrickNet, this also increases collision-free builds from 30.9\% to 47.4\%.

\paragraph{The Importance of Parent Selection.}
On BrickNet, 99.9\% of the invalid parts in B3 are attached to parents with no valid candidate placements (Sec.~\ref{app:fallback} in the supplementary).
In these cases, decoding falls back to grammar constraints alone, often resulting in invalid placements.
The parent veto (B4) prevents the model from selecting such parents, reducing invalid parts from 7.4\% to 0.5\% and increasing collision-free builds from 47.4\% to 90.8\%, at only 1.26$\times$ the per-token cost of unconstrained decoding (Table~\ref{tab:bn_operator_cost} in the supplementary).
On ICLF, the parent veto (I3) similarly reduces invalid parts from 12.1\% to 5.8\% and increases collision-free builds from 31.4\% to 65.2\%.

\subsection{Constraint Strength Ablation}
\label{sec:hard_vs_soft}

Our formulation controls the strength of geometric constraints through $\beta$ (Eq.~\ref{eq:soft_reference}). 
With a finite $\beta$, invalid placements are penalized rather than completely masked out, allowing the model to override the geometric constraints when it is confident enough.
This flexibility may be useful when the geometric oracle does not include the placement intended by the model.
We investigate whether such overrides improve visual fidelity by varying $\beta$ while keeping the parent veto disabled.
At $\beta=0$, only grammar and catalog constraints are applied (B1/I1). Increasing $\beta$ penalizes invalid placements more strongly, until $\beta\to\infty$ imposes hard placement constraints (B3/I2).

Figure~\ref{fig:soft_pareto} shows the trade-off between geometric validity and visual fidelity for both models.
On BrickNet-14B, increasing $\beta$ substantially reduces invalid placements, while DINO similarity remains nearly unchanged. Thus, soft constraints provide little benefit over the hard mask.
On ICLF, the penalty removes less than half of the invalid parts that the hard mask removes: at $\beta{=}8$, 24.4\% of parts are invalid against 12.1\% under the mask, while DINO similarity varies by at most 0.005 across the sweep. The VQA score rises slightly with $\beta$ (0.279 at $\beta{=}0$ to 0.309 at $\beta{=}8$, against 0.292 under the mask), at twice the share of invalid parts.
Overall, the hard mask provides the most favorable trade-off for both models.

To understand why soft constraints are less effective, we analyze 256 BrickNet-14B builds at each $\beta$ and examine cases where the model overrides the geometric constraints (Sec.~\ref{app:soft_override} in the supplementary).
Even at $\beta{=}1$, overrides put nearly all the probability outside $A_i$, so they are confident predictions, not sampling noise.
However, most overrides (65--73\%) select placements included in the oracle but rejected by the collision checker, and approximately 70\% of all overrides result in collisions.
These findings suggest that the model's confidence is not a reliable indicator of geometric validity, motivating our use of hard constraints.

\section{Limitations}
\label{sec:limitations}
\begin{figure}[t]
\centering
\includegraphics[width=\columnwidth]{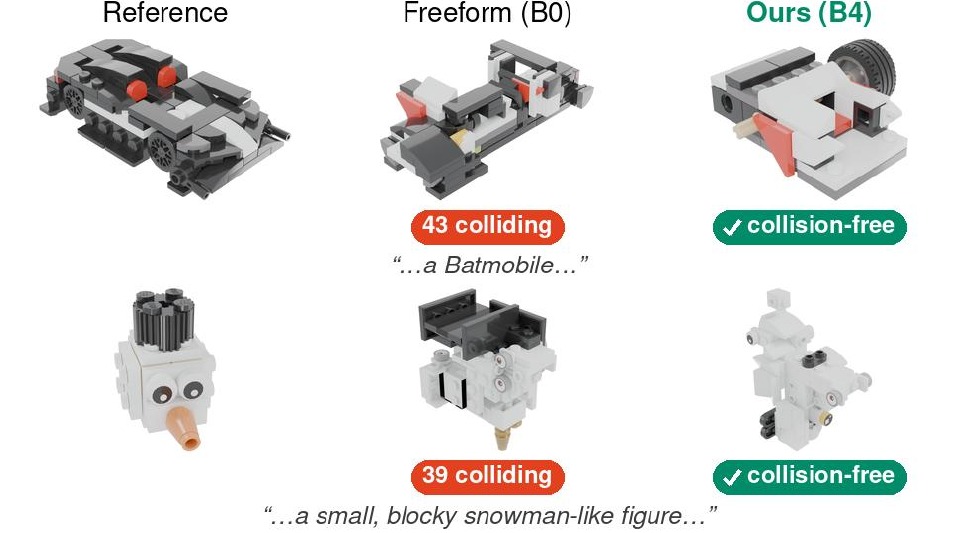}
\caption{Limitations. When the base model cannot build the captioned object (left), our method makes the build collision-free (right) but no more recognizable than unconstrained decoding (middle): we constrain how parts attach, not which parts the model chooses, nor by the appearance of a placement}
\label{fig:limitations}
\end{figure}
Our validity is geometric: we check collisions and attachments, not stability~\cite{pun2025brickgpt}, so a valid build can still fall over.
Another limitation is that our method relies on a partially relational format where each part names its parent, and part type is not constrained.
This means that not every output token is constrained, and the model may commit to a part type that cannot be added validly to the build.
Finally, our method inherits the visual quality of the base model, which is limited: many of its builds resemble their captions only loosely (Fig.~\ref{fig:limitations}),
and since we constrain only how parts attach, our builds look no better.

\section{Conclusion}
We presented a constrained decoding method that enforces geometric validity during decoding, without retraining the model.
Instead of enumerating every placement, we let the model choose a part and its parent, list how the two can attach with a precomputed oracle, filter that list against the current build, and bias the next-token distribution towards valid attachments;
a parent veto avoids parents that have no valid attachment left.
The gains hold under ColSAT, a checker the decoder never queries, and on ICLF, a format without connector annotations, where the constraints cut invalid parts from 27\% to 6\%, at the cost of a lower VQA score.
This work shows that incorporating closed-form procedures into autoregressive generation can improve a pretrained LLM's performance on tasks it struggles with, with little harmful policy drift.
Applying the approach to other domains where LLMs violate easily verifiable constraints, such as robotics and simulation remains future work. 

\clearpage
\appendix
\renewcommand{\thesection}{\Alph{section}}
\setcounter{figure}{0}\setcounter{table}{0}\setcounter{equation}{0}
\renewcommand{\thefigure}{S\arabic{figure}}\renewcommand{\theHfigure}{S\arabic{figure}}
\renewcommand{\thetable}{S\arabic{table}}\renewcommand{\theHtable}{S\arabic{table}}
\renewcommand{\theequation}{S\arabic{equation}}\renewcommand{\theHequation}{S\arabic{equation}}
\maketitlesupplementary
\section{Implicit Connectivity LDraw Format (ICLF)}
\label{app:format_examples}
\begin{figure*}[t]
\centering
\includegraphics[width=\textwidth]{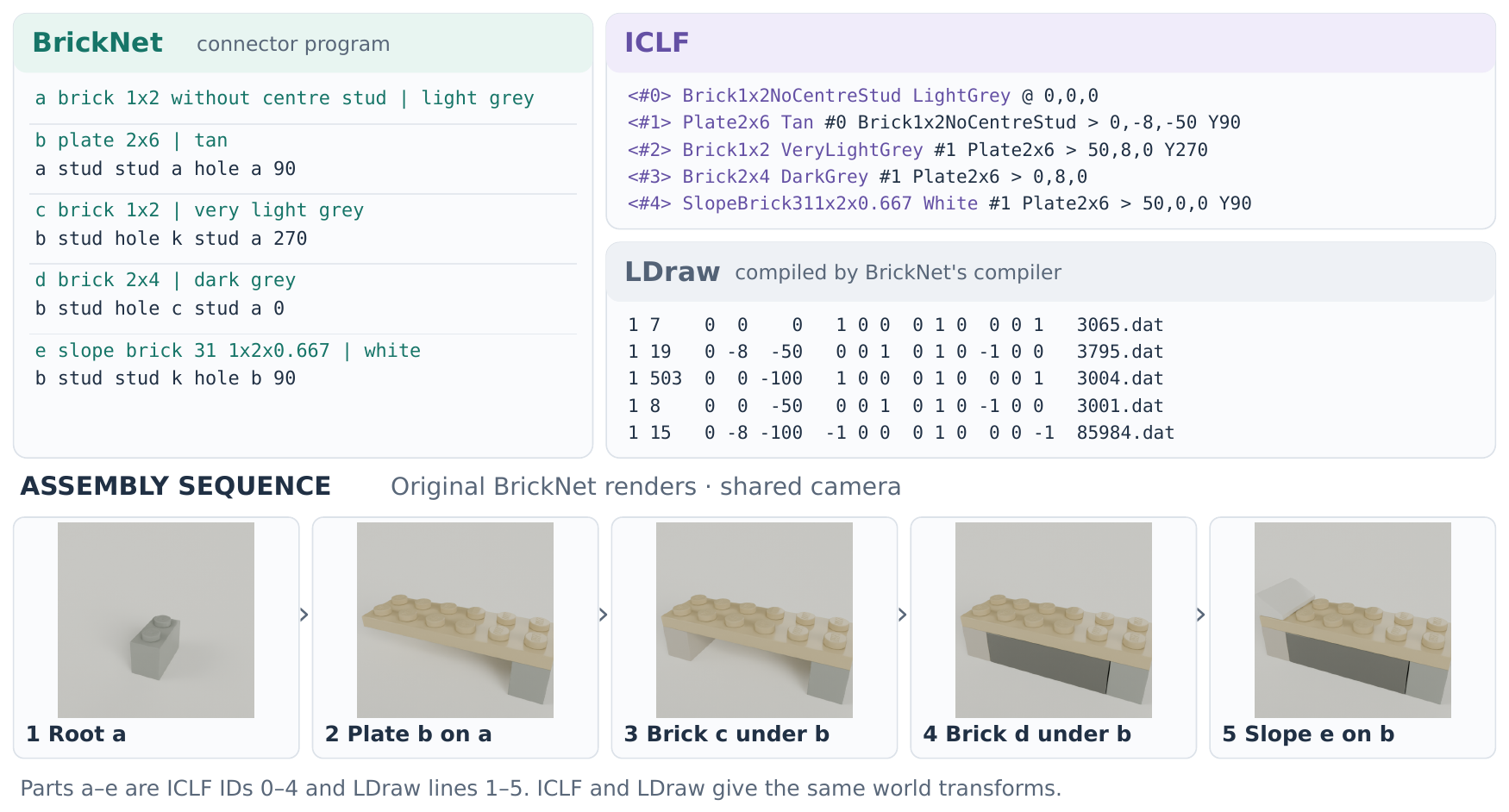}  %
\caption{
One build in three forms: the BrickNet connector program, the ICLF records of the compiled LDraw file, and the LDraw file itself. 
BrickNet writes two lines per part (the part, then its attachment edge), ICLF and LDraw write one line per part. 
Parts appear in the same order in all three forms.
}
\label{fig:appendix_one_build}
\end{figure*}
\subsection{Record Format and Rotation Encoding}
\label{app:rldr_details}
An ICLF part line contains its index in the build, part-type name, color, reference, translation, and orientation. 
A relative parent reference includes the index of an earlier part and its part-type name. 
If the parent has world pose $(R_p,t_p)$ and the record specifies $(R_r,d)$, the child pose is $R_c=R_pR_r$, $t_c=t_p+R_pd$. An absolute record instead specifies $(R_c,t_c)$ directly. Translation is in LDraw~\cite{ldraw} units. %
Orientation encoding is detailed below.

\paragraph{Part fields.}
Part and color fields use readable natural language names rather than LDraw's numerical identifiers. Each record has an explicit brick index; a relative reference repeats the parent's part-type name as well as its index (Fig.~\ref{fig:appendix_one_build}). The record specifies a relative offset and rotation directly, so connector compatibility must be inferred from the resulting placement rather than from an explicit connector-instance pair. This geometric interface is shared across connector families.

\paragraph{Orientation encoding.}
\label{app:rldr_orientation}
The orientation field represents the parent-relative rotation $R_r$ for a referenced brick, or the world rotation $R_c$ for an absolute \texttt{@} record. 
The compact vocabulary covers the 24 proper rotations that align the brick's local axes with the reference axes. 
It includes the identity transform \texttt{I}, single-axis labels such as \texttt{Z90}, and compound labels such as \texttt{X90.Y90}. 
Angles are in degrees. 
A compound label is decoded as an ordered matrix product: \texttt{X90.Y90} denotes $R_x(90^\circ)R_y(90^\circ)$, with the rightmost rotation acting first on a column vector. 
An omitted orientation field is parsed as identity; the converter omits explicit \texttt{I}.

Other orientations use axis--angle notation \texttt{R:ax,ay,az,deg}, whose axis is normalized during decoding, or an explicit matrix \texttt{M:} followed by nine comma-separated entries in row-major order. Matrix notation represents
\begin{equation}
R=\begin{bmatrix}
m_{11}&m_{12}&m_{13}\\
m_{21}&m_{22}&m_{23}\\
m_{31}&m_{32}&m_{33}
\end{bmatrix}.
\end{equation}

\paragraph{Reference kinds.}
A snap reference (\texttt{>}) names an already placed part that the new part connects to in the LDraw connector graph (LDCad shadow metadata~\cite{ldcadshadow}); a proximity reference (\texttt{\textasciitilde}) names the nearest placed part for an attachment that the build and BrickNet's connection graph contain but LDCad's connector metadata does not confirm, so such parts keep a short, local offset; an absolute record (\texttt{@}) has no parent and starts a new root. A generated \texttt{>} is the model's prediction, not evidence of a valid connection.
During constrained decoding the reference kind is chosen by the decoder: it offers the oracle's connector-based candidates (\texttt{>}) that pass the collision check, otherwise the oracle's proximity placements (\texttt{\textasciitilde}) observed in the training builds, under the same check; if neither survives, it writes \texttt{\textasciitilde} and the offset is decoded with the grammar alone.

\subsection{Prefix-Tree Shape}
\label{app:rldr_trie}
The two formats give the candidate prefix tree (Sec.~\ref{sec:token_constraints} in the main paper) different shapes (Table~\ref{tab:trie_shape}).
We sample 200 part-type pairs from BrickNet's~\cite{kulits2026bricknet} validation builds, weighted by how often they occur, and compare the same attachments written in both syntaxes.
For each pair we take BrickNet's enumerated candidates, merge strings that describe the same physical placement (BrickNet can name one placement with several connector pairs, about two strings per placement),
write each placement once in BrickNet and once in ICLF syntax, and tokenize it with the corresponding model's tokenizer.
On identical placements, an ICLF attachment takes 12.7 tokens against 9.6, and its tree is deeper (median depth 14 against 10).
The branching also moves: BrickNet's tree branches mostly at the parent-connector and child-connector tokens, 
while ICLF's branches thinly at nearly every digit of the offset and the rotation.
\begin{table}[t]
\centering\small
\setlength{\tabcolsep}{5pt}
\begin{tabular}{lrr}
\toprule
 & BrickNet & ICLF\\
\midrule
Candidates per pair, median & 175 & 175\\
Tokens per attachment, mean & 9.6 & 12.7\\
Tree depth, median & 10 & 14\\
Tree nodes, median & 753 & 947\\
Mean branching, two widest levels & 15.8, 4.7 & 5.7, 2.0\\
Mean branching, other levels & $<$3 & $\le$1.8\\
\bottomrule
\end{tabular}
\caption{
  Shape of the candidate prefix tree before collision filtering, over 200 occurrence-weighted part-type pairs from BrickNet's validation builds. 
  Both columns hold the same placements (strings that describe one physical placement merged), written in each format and tokenized with that model's tokenizer. Branching: the mean number of children of a node at one tree level, averaged over pairs; BrickNet's two widest levels are the parent-connector and child-connector tokens, ICLF's are its first two tokens.
}
\label{tab:trie_shape}
\end{table}

\subsection{Model Training}
\label{app:rldr_training}
Our ICLF model fine-tunes Qwen3.6-27B~\cite{qwen36} with a rank-512 LoRA~\cite{hu2022lora} ($\alpha{=}512$, no dropout) on all linear projections.
We train on BrickNet's training builds converted to ICLF, and compute the loss on the assistant response only.
Minifigure parts are removed from every training build (parts the LDraw catalog files as minifigure parts or whose name begins with ``Minifig''), so the model does not emit minifigures.
For the ICLF evaluation we therefore drop the BN-val-512 captions that ask for one: a caption is dropped if ``minifig'', ``minifigure'' or ``figure'' occurs in its first 200 characters, which removes 85 captions and leaves 427.
We use AdamW~\cite{loshchilov2019adamw} ($\beta{=}(0.9, 0.999)$, no weight decay, gradient clipping at 1.0) in bf16 with a global batch of 48 conversations of up to 32,768 tokens on four B200 GPUs.
The first stage warms up over 50 steps to a learning rate of $2\times10^{-4}$ with cosine decay over a 2,800-step schedule, and we stop it at step 2,600.
A warm restart continues from that adapter with a peak learning rate of $5\times10^{-5}$ (50 warmup steps, cosine decay over 3,000 steps); we take its step 2,000 (4,600 steps in total) and merge the adapter into the base weights for decoding.
\subsection{One Build in LDraw, BrickNet and ICLF}
\label{app:one_build}
Figure~\ref{fig:appendix_one_build} shows one five-part build in the three forms used in this paper. 
The ICLF records encode the LDraw file, with the parts in the same order.

\paragraph{LDraw file.}
\label{app:ldraw_example}
The LDraw panel of Fig.~\ref{fig:appendix_one_build} is the output of BrickNet's own compiler for the BrickNet program in the same figure.
Each LDraw line has type~\texttt{1} (a part reference), followed by the color code, the position $x\ y\ z$ in LDraw units (one stud pitch is 20 units and the $y$ axis points down), the nine entries of the rotation matrix row by row, and the part file.
For example, the second line places a tan (\texttt{19}) $2{\times}6$ plate (\texttt{3795.dat}) 8 units above the root, rotated by $90^\circ$ about the vertical axis.
Nothing in the file states which parts connect.
\section{Collision Checkers}
\subsection{BrickNet Mesh Checker and GPU Port}
\label{app:mesh_checker}
BrickNet's collision test replaces each part by its collision mesh shrunk (inset) by 0.25~LDU and declares two parts colliding when a triangle of one crosses a triangle of the other; coincident duplicate parts also collide. 
The released scorer applies this test with MeshLib~\cite{meshlib} after a spatial broad phase. 
A part joined to its parent by a fixed mount is meant to interpenetrate it, so BrickNet's scorer skips that one pair; every other pair is checked. 
Our decoder uses a Warp~\cite{warp2022} GPU port of the same test (per-part bounding-volume hierarchies, candidate poses batched against a shared scene, incremental per-rollout scenes).
On 6,000 BrickNet-14B output builds for BN-val-512 (three decoder configurations), the port and the released scorer agree on 70,263 colliding pairs; the port reports 29 additional pairs and misses none, and the two flag the same builds.
Our decoders apply the same exemption.
ICLF outputs are converted to LDraw and scored with the same released collision test, with parts checked in record order; 
BrickNet's LDraw parser recovers connections from the geometry, and fixed-mount pairs are exempted in the same way.

\label{app:validation_details}

\label{app:bricknet_validation}

\subsection{ColSAT: Oriented Collision Boxes}
\label{app:colsat_boxes}
\paragraph{Separating-axis box test.}
\label{app:colsat_axes}
ColSAT applies the standard separating-axis test~\cite{eberlyseparatingaxes} to every pair of Studio~\cite{bricklinkstudio} collision boxes; 
touching boxes do not collide.
\label{app:validation_policies}
Table~\ref{tab:bn_operator_ablation} in the main paper counts only the later-placed part of each remaining colliding pair.
\section{Geometric Oracle Coverage}
\label{sec:oracle_coverage}
We compare candidate sources on 32,248 directed attachment occurrences from the BrickNet validation set, spanning 5,524 ordered part-type pairs and 1,137 part types. 
An attachment counts as covered when its exact connector pair and pose are among the oracle's candidates.
The oracle is a single set of placements shared by both formats: an ICLF attachment and a BrickNet attachment are the same placement written in different syntax, so the coverage below holds for both.

Enumeration samples each free pose variable of a connector pair on a fixed grid: rotations about the connector axis in $90^\circ$ steps and slides along it of $\{-20,0,20\}$\, LDU, with at most 65,536 lines stored per part-type pair.

\begin{table}[t]
\centering\small
\setlength{\tabcolsep}{3pt}
\begin{tabular}{lrrrr}
\toprule
Family & Edges & Empirical & Enumerated & Union\\
\midrule
Stud & 29,455 & 78.16 & 97.68 & 97.98\\
Axle & 1,641 & 63.74 & 15.30 & 67.58\\
Hinge & 544 & 90.99 & 43.20 & 92.46\\
Ball & 60 & 78.33 & 73.33 & 85.00\\
Fixed & 548 & 95.99 & 100.00 & 100.00\\
\midrule
All edges & 32,248 & 77.94 & 92.56 & 96.35\\
Pair mean & -- & 70.65 & 87.06 & 93.34\\
\bottomrule
\end{tabular}
\caption{
Oracle coverage (\%) of BrickNet validation builds. 
The last row averages within-pair coverage across 5,524 pairs. 
}
\label{tab:oracle_coverage}
\end{table}

The enumerated part of the oracle is precomputed once and stored per part-type pair, built in 538\,s on 128 CPU cores. 
Enumeration supplies 5,936 occurrences absent from the oracle's empirical attachments (those observed in BrickNet's training builds), while empirical candidates supply 1,222 absent from enumeration; their union covers 96.35\%. 
Enumeration alone is nearly complete for studs (97.68\%) but covers only 15.30\% of axle and 43.20\% of hinge attachments, whose off-grid placements the oracle's empirical attachments supply. 
Five pairs exceed the 65,536-attachment cap; their 488 occurrences remain in the denominator and account for 488 of the union's 1,177 misses.

\section{Additional Results}
\label{app:operator_cost}
\paragraph{Decoding setup.}
Both models are served with vLLM~\cite{kwon2023vllm}, and every decoder samples four outputs per caption with the same four seeds.
BrickNet-14B is prompted with the caption and sampled at temperature 1.0 with top-$p$ 0.95 and top-$k$ 20, for at most 3,800 generated tokens, within a 4,096-token serving context that includes the prompt.
Our ICLF model is prompted with its training system prompt followed by the BrickNet caption and sampled at temperature 1.0, for at most 8,192 generated tokens, in bf16 on one B200 GPU per shard.
Decoding sets no cap on the number of parts.

\subsection{Decoding Cost and Rejection Budgets}
Table~\ref{tab:bn_operator_cost} gives the parse yield and decoding cost of the final mesh-checker decoders on both models.
\begin{table}[t]
\centering\footnotesize
\setlength{\tabcolsep}{3pt}
\begin{tabular}{lrrr}
\toprule
Arm & Parse & CF & Time/tok\\
 & (\%) $\uparrow$ & (\%) $\uparrow$ & ($\times$ freeform) $\downarrow$\\
\midrule
\multicolumn{4}{@{}l}{\emph{BrickNet-14B, BN-val-512}}\\
B0 freeform & 94.2 & 30.9 & 1.00\\
B1 + grammar \& catalog & 99.5 & 29.1 & --\\
B4 + parent veto & 99.4 & 90.8 & 1.26\\
Rejection, $N{=}2$ & 100.0 & 49.6 & 1.65\\
Rejection, $N{=}4$ & 100.0 & 64.3 & 2.01\\
Rejection, $N{=}8$ & 100.0 & 74.3 & 2.14\\
\midrule
\multicolumn{4}{@{}l}{\emph{ICLF, 427 BN-val captions}}\\
I0 freeform & 98.7 & 14.6 & 1.00\\
I1 + grammar \& catalog & 98.8 & 7.4 & --\\
I3 + parent veto & 98.8 & 65.2 & 1.56\\
Rejection, $N{=}2$ & 99.8 & 20.8 & 1.57\\
Rejection, $N{=}4$ & 99.8 & 26.5 & 1.99\\
Rejection, $N{=}8$ & 99.8 & 32.8 & 2.47\\
\bottomrule
\end{tabular}
\caption{Parse yield and decoding cost of the final mesh-checker decoders on both models. Parse: \% of the attempts (2048 for BrickNet-14B, 1708 for ICLF) that are parsable; CF: collision-free attempts under the BrickNet checker (Table~\ref{tab:bn_operator_ablation} in the main paper). Time/tok: wall time per token kept in the final build (rejection counts its discarded tries), relative to freeform decoding on one B200 GPU (96 builds per configuration).}
\label{tab:bn_operator_cost}
\end{table}

Table~\ref{tab:rejection} compares the parent-veto decoder with per-part rejection sampling at every budget on both models (Sec.~\ref{sec:rejection} and Fig.~\ref{fig:rejection_cost} in the main paper).
\begin{table}[t]
\centering\small
\setlength{\tabcolsep}{4pt}
\begin{tabular}{lrrrr}
\toprule
 & CF & \multicolumn{2}{c}{Part inv. (\%) $\downarrow$} & Time/tok\\
\cmidrule(lr){3-4}
Arm & (\%) $\uparrow$ & BN & ColSAT & ($\times$ freeform) $\downarrow$\\
\midrule
\multicolumn{5}{@{}l}{\emph{BrickNet-14B, BN-val-512}}\\
Freeform (B0) & 30.9 & 23.7 & 23.6 & 1.00\\
Rejection, $N{=}2$ & 49.6 & 10.5 & 13.5 & 1.65\\
Rejection, $N{=}4$ & 64.3 & 5.3 & 9.1 & 2.01\\
Rejection, $N{=}8$ & 74.3 & 2.7 & 6.9 & 2.14\\
B4 parent veto (ours) & \textbf{90.8} & \textbf{0.5} & \textbf{4.2} & 1.26\\
\midrule
\multicolumn{5}{@{}l}{\emph{ICLF, 427 BN-val captions}}\\
Freeform (I0) & 14.6 & 27.3 & 22.2 & 1.00\\
Rejection, $N{=}2$ & 20.8 & 19.6 & 16.4 & 1.57\\
Rejection, $N{=}4$ & 26.5 & 14.9 & 13.2 & 1.99\\
Rejection, $N{=}8$ & 32.8 & 11.1 & 10.5 & 2.47\\
I3 parent veto (ours) & \textbf{65.2} & \textbf{5.8} & \textbf{6.0} & 1.56\\
\bottomrule
\end{tabular}
\caption{Constrained decoding against per-part rejection sampling at every budget: the data of Fig.~\ref{fig:rejection_cost} in the main paper, four samples per caption. CF: collision-free builds (BrickNet checker); Part inv.: \% of parts that are the later-placed part of a colliding pair. Rejection re-draws a part while the collision checker flags it or it does not parse, up to $N$ tries. Time/tok: wall time per token kept in the final build (rejection counts its discarded tries), relative to freeform decoding on one B200 GPU (96 builds per configuration).}
\label{tab:rejection}
\end{table}

\subsection{Where the Remaining Invalid Parts Come From}
\label{app:fallback}
We mark each attachment whose chosen parent had no surviving candidate, so that decoding fell back to the grammar (Sec.~\ref{sec:token_constraints} in the main paper), and compare these with the invalid parts under the BrickNet checker, for BrickNet-14B on BN-val-512.
In B3, 7.6\% of the attachments fall back, and they account for 99.93\% of the invalid parts (5{,}408 of 5{,}412); every attachment that falls back is invalid.
With the parent veto (B4), fallbacks drop to 0.55\% of the attachments and still account for 386 of the 388 invalid parts.
\section{Soft Guidance}
\label{app:soft_intervals}

\subsection{Setup}
\label{app:soft_setup}
Fig.~\ref{fig:soft_pareto} in the main paper gives the $\beta$ sweeps of both models under the soft penalty of Eq.~\eqref{eq:soft_reference} in the main paper, with the parent veto disabled and decoded with the mesh checker; $\beta{=}0$ applies only the grammar and catalog constraints, and $\beta\to\infty$ is the hard mask.
We say the model \emph{overrides} the penalty at an attachment when $A_i$ is non-empty and it samples a token outside $A_i$, so the part it commits is not one of the candidates that passed the collision check.

\subsection{Where the Model Overrides the Penalty}
\label{app:soft_override}
We force the saved outputs of 256 BrickNet-14B builds per $\beta$ (captions 0--127, two samples each) back through the soft decoder; it reproduces every output token for token, and we record at each attachment whether a sampled token left $A_i$.
Table~\ref{tab:soft_override} summarizes the overrides.
They are confident: at the token where the model leaves $A_i$, its median probability outside $A_i$ before the penalty is above 0.99 at every $\beta$.
Most overrides (65--73\%) commit an oracle placement that the collision test rejected, and 93--95\% of these collide.
The others commit a placement the oracle does not contain; most of these are valid (6--26\% invalid), consistent with the oracle missing some valid placements (Sec.~\ref{sec:oracle_coverage}).
About three quarters of the overrides leave $A_i$ at the connector tokens, before any pose number is written, and 82--86\% of these are invalid, against 17--38\% for overrides at the pose numbers.
\begin{table}[t]
\centering\footnotesize
\setlength{\tabcolsep}{3pt}
\begin{tabular}{lcccccc}
\toprule
 & Overrides & Invalid & In oracle & \multicolumn{2}{c}{Invalid (\%)} & Prob.\ outside \\
\cmidrule(lr){5-6}
$\beta$ & (\% att.) & (\%) & (\%) & in oracle & off oracle & $A_i$ (median)\\
\midrule
1 & 18.1 & 69.8 & 73.2 & 93.1 & 6.5 & 0.993\\
2 & 16.4 & 71.0 & 71.4 & 94.7 & 11.9 & 0.998\\
4 & 11.2 & 72.0 & 72.1 & 95.1 & 12.3 & 1.000\\
8 & 4.7 & 70.4 & 65.3 & 94.1 & 25.7 & 1.000\\
\bottomrule
\end{tabular}
\caption{Overrides of the soft penalty, BrickNet-14B, 256 replayed builds per $\beta$.
Overrides: share of all attachments at which a sampled token left a non-empty $A_i$ (attachments with an empty $A_i$, 7--10\%, are not counted).
Invalid: the committed part collides with an earlier part under BrickNet's mesh checker (Sec.~\ref{app:bricknet_validation}).
In oracle: overrides whose committed placement is an oracle placement rejected by the collision test.
Prob.\ outside $A_i$: the model's probability mass outside $A_i$ at the token where it left, before the penalty.}
\label{tab:soft_override}
\end{table}

\section{Qualitative Examples}
\label{app:qualitative_ladder}
\begin{figure*}[p]\centering
\includegraphics[width=\textwidth]{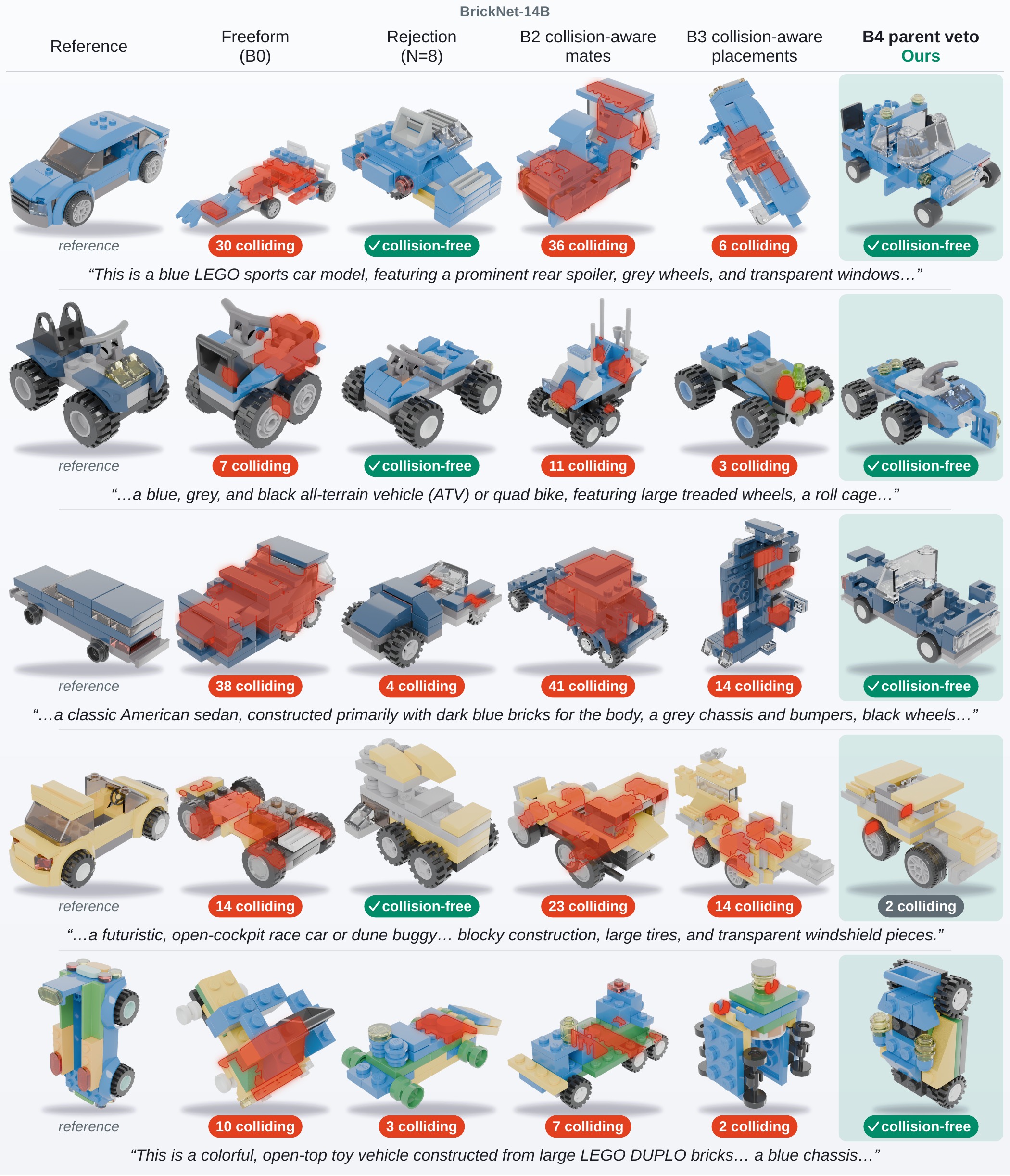}
\caption{Qualitative examples (1/2): BrickNet-14B on BN-val-512 captions. Columns follow Table~\ref{tab:bn_operator_ablation} in the main paper without B1. ``Colliding'' counts the parts in a collision under the BrickNet mesh checker.}
\label{fig:supp_qual_bn_1}
\end{figure*}

\begin{figure*}[p]\centering
\includegraphics[width=\textwidth]{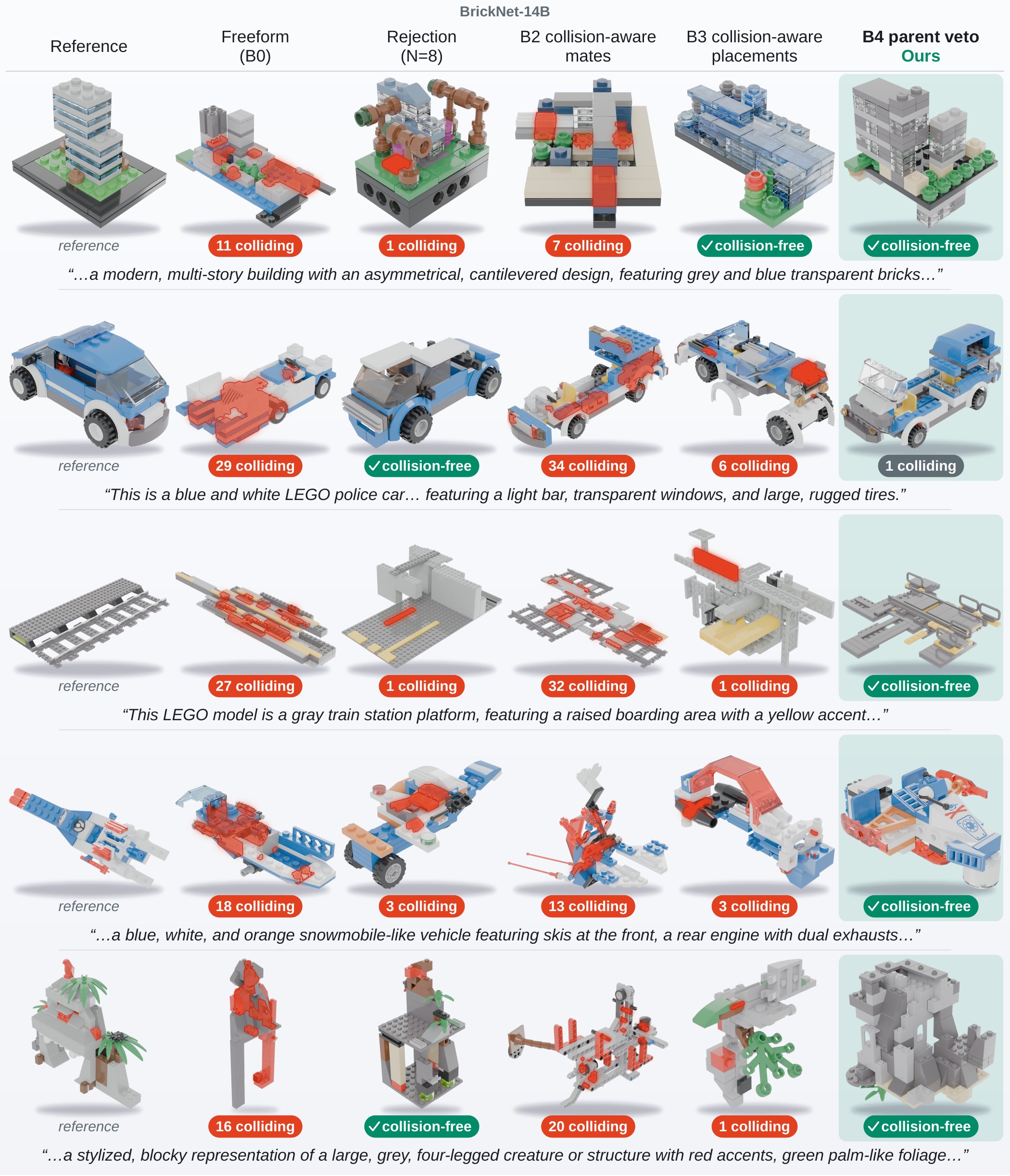}
\caption{Qualitative examples (2/2): BrickNet-14B on BN-val-512 captions. Columns follow Table~\ref{tab:bn_operator_ablation} in the main paper without B1. ``Colliding'' counts the parts in a collision under the BrickNet mesh checker.}
\label{fig:supp_qual_bn_2}
\end{figure*}

\begin{figure*}[p]\centering
\includegraphics[width=\textwidth]{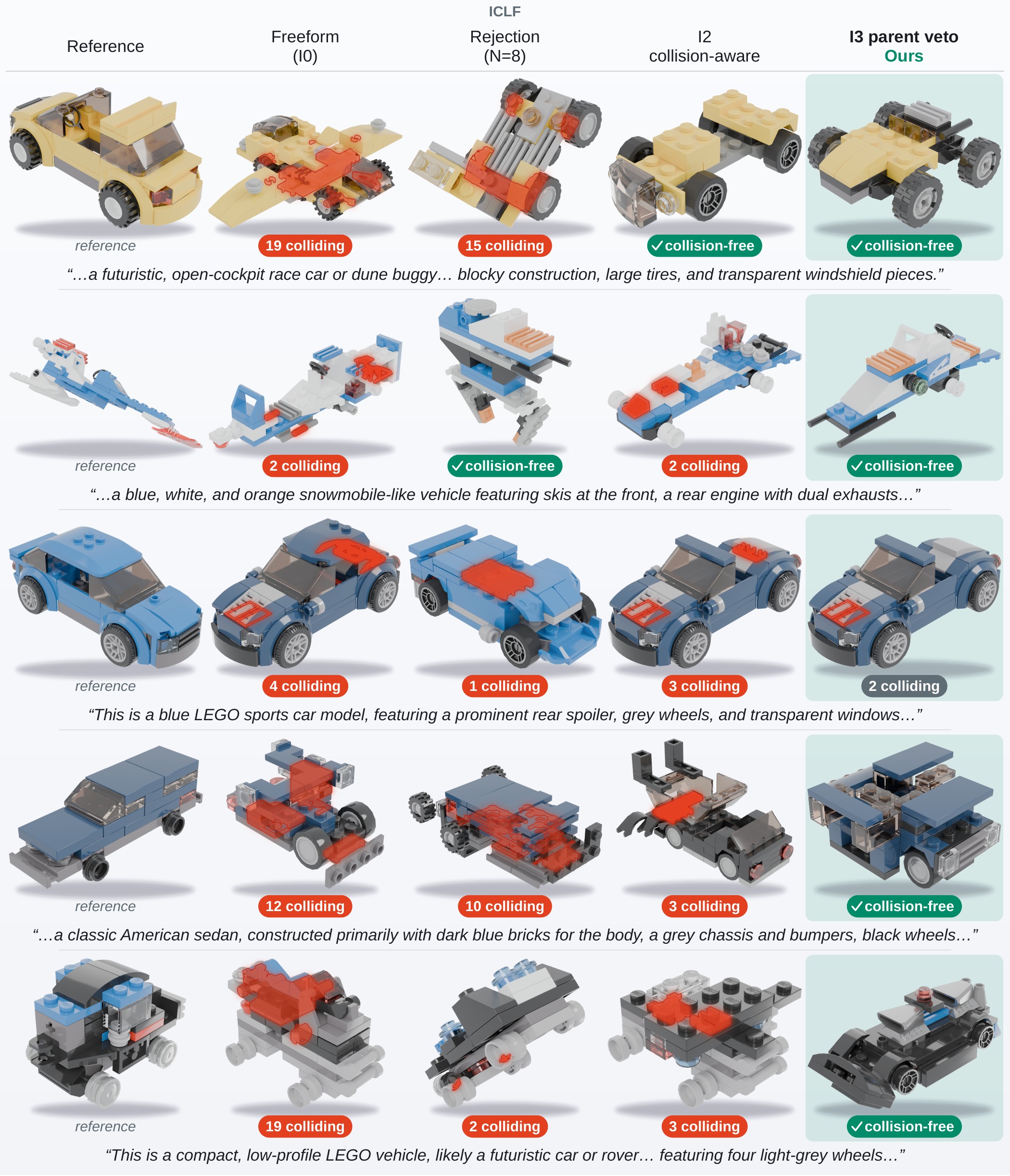}
\caption{Qualitative examples (1/2): our ICLF model. Columns follow Table~\ref{tab:bn_operator_ablation} in the main paper. ``Colliding'' counts the parts in a collision under the BrickNet mesh checker.}
\label{fig:supp_qual_iclf_1}
\end{figure*}

\begin{figure*}[p]\centering
\includegraphics[width=\textwidth]{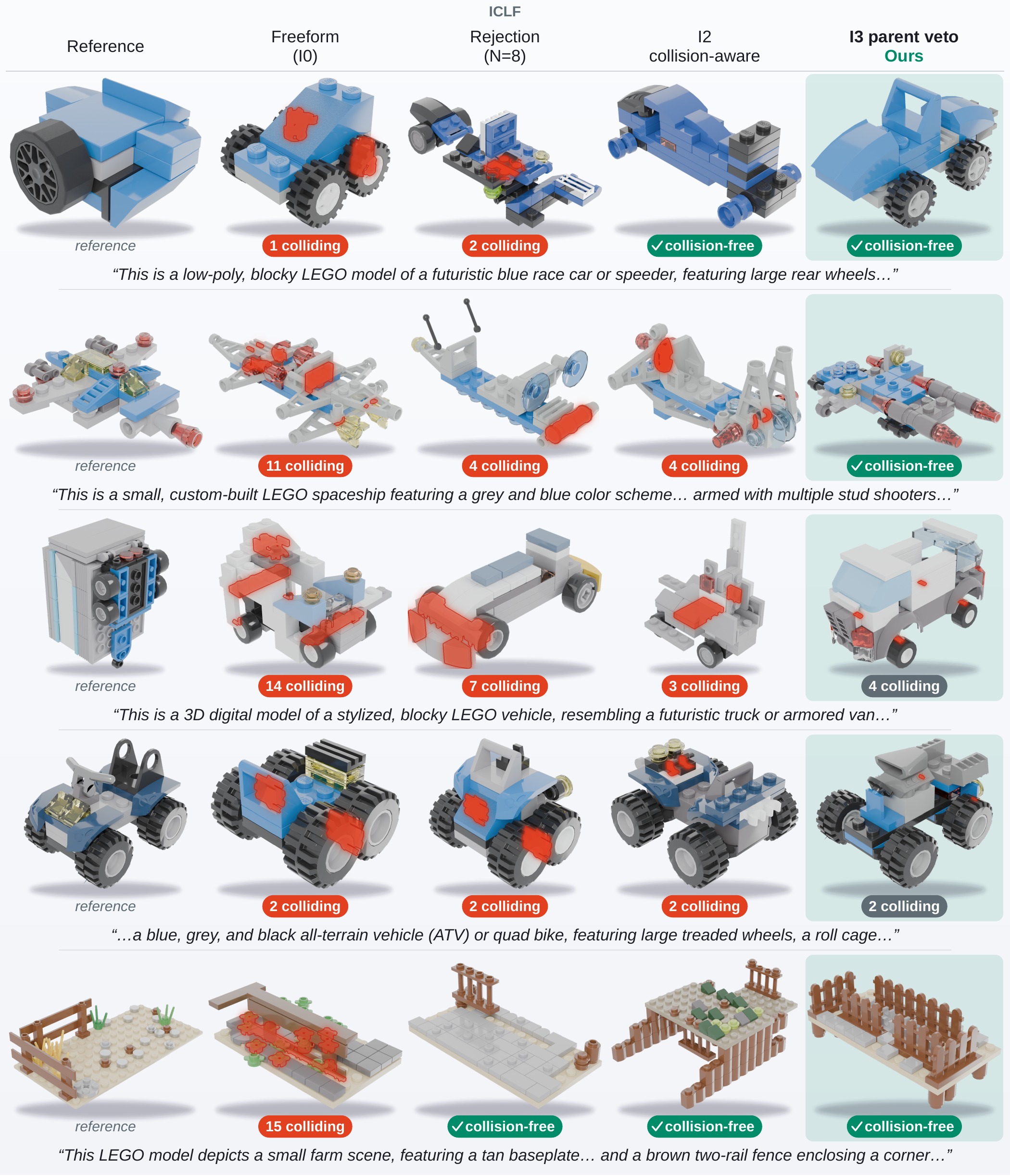}
\caption{Qualitative examples (2/2): our ICLF model. Columns follow Table~\ref{tab:bn_operator_ablation} in the main paper. ``Colliding'' counts the parts in a collision under the BrickNet mesh checker.}
\label{fig:supp_qual_iclf_2}
\end{figure*}

Figures~\ref{fig:supp_qual_bn_1}--\ref{fig:supp_qual_bn_2} show ten BN-val-512 captions on BrickNet-14B and Figs.~\ref{fig:supp_qual_iclf_1}--\ref{fig:supp_qual_iclf_2} ten captions on our ICLF model, with the decoders of Table~\ref{tab:bn_operator_ablation} in the main paper.

\end{document}